\documentclass[wcp]{jmlr}

\usepackage{longtable}

\usepackage{booktabs}
\usepackage{graphicx}
\usepackage{xcolor}
\usepackage{multirow}
\usepackage{colortbl}
\usepackage{ifpdf}
\usepackage[utf8]{inputenc}
\usepackage{graphicx}

\usepackage{multirow}
\usepackage{multicol}
\usepackage{booktabs}
\usepackage{makecell}
\usepackage{siunitx}
\usepackage{tcolorbox}

\usepackage{mdwmath}
\usepackage{mdwtab}

\usepackage{natbib}

\usepackage{lineno}

\makeatletter
\let\Ginclude@graphics\@org@Ginclude@graphics 
\makeatother

\jmlrvolume{222}
\jmlryear{2023}
\jmlrworkshop{ACML 2023}

\title[Temporal Shift]{Temporal Shift - Multi-Objective Loss Function for Improved Anomaly Fall Detection}

  \author{\Name{Stefan Denkovski} \Email{stefan.denkovski@mail.utoronto.ca}\\
   \Name{Shehroz S. Khan} \Email{shehroz.khan@uhn.ca}\\
   \Name{Alex  Mihailidis} \Email{alex.mihailidis@utoronto.ca}\\
   \addr KITE Research Institute – University Health Network, Toronto, Canada \\
   \addr Institute of Biomedical Engineering, University of Toronto, Toronto, Canada}

\editors{Berrin Yan{\i}ko\u{g}lu and Wray Buntine}

\begin{document}
\maketitle

\begin{abstract}

Falls are a major cause of injuries and deaths among older adults worldwide. Accurate fall detection can help reduce potential injuries and additional health complications. Different types of video modalities can be used in a home setting to detect falls, including RGB, Infrared, and Thermal cameras. Anomaly detection frameworks using autoencoders and their variants can be used for fall detection due to the data imbalance that arises from the rarity and diversity of falls. However, the use of reconstruction error in autoencoders can limit the application of networks' structures that propagate information. In this paper, we propose a new multi-objective loss function called Temporal Shift, which aims to predict both future and reconstructed frames within a window of sequential frames. The proposed loss function is evaluated on a semi-naturalistic fall detection dataset containing multiple camera modalities. The autoencoders were trained on normal activities of daily living (ADL) performed by older adults and tested on ADLs and falls performed by young adults. Temporal shift shows significant improvement to a baseline 3D Convolutional autoencoder, an attention U-Net CAE, and a multi-modal neural network. The greatest improvement was observed in an attention U-Net model improving by $0.20$ AUC ROC for a single camera when compared to reconstruction alone. With significant improvement across different models, this approach has the potential to be widely adopted and improve anomaly detection capabilities in other settings besides fall detection. 

\end{abstract}


\section{Introduction}


The number of older adults is growing, as both a ratio and absolute numbers in populations, creating a tremendous challenge for health care systems \cite{fuster2017changing}. Falls are a 
leading cause of injury and death in older adults \cite{KramarowEChenLHHedegaardH2015}. To reduce the risk of complications, fall detection is important as it allows for early intervention and reduce additional health complications \cite{Stinchcombe2014, Rubenstein2002}. It also enables older adults to live independently by ensuring that they receive assistance when needed. Video or computer vision detection systems are preferred for fall detection, as they require no user input, are non-invasive, and can work in a variety of environments, including the home \cite{Igual2013}. However, fall detection is a challenging problem, both from a predictive modeling perspective and practical considerations, such as privacy and false alarm rates \cite{khan2014one, Khan2017}.

Predictive modeling for fall detection is challenging due to the rarity and diversity of fall events, making it difficult to create a well-defined class for falls or to capture all possible variations of falls in a dataset \cite{Khan2017}. Previous research shows that natural falls occur only on 0.3\% to  1.6\% of days \cite{stone2014fall, debard2012camera}, and even then, each fall may last only a few seconds, resulting in very highly skewed datasets \cite{Khan2017}. Therefore, even long-term studies may not contain enough fall data to build a robust supervised classifier \cite{Khan2016}. Additionally, each rare fall event can vary greatly from one another, making it even difficult to capture possible variations \cite{Khan2016, charfi2012definition}. Finally, practical considerations, such as low false alarm rates and privacy, need to be taken into account when developing fall detection systems, as they can significantly affect the system's usefulness and adoption.

This paper introduces a novel multi-objective loss function, Temporal Shift, which significantly enhances the optimization of video anomaly detection frameworks. Temporal Shift is comprised of reconstruction loss and prediction loss and is applied to various convolutional autoencoder network structures. The networks are evaluated on the Multi-Visual Modality Fall Detection Dataset (MUVIM) \cite{denkovski2022multi}, demonstrating that Temporal Shift enables:

\begin{itemize}
  \item Improved performance of 3D Convolutional autoencoder (3DCAE) for fall detection.
  \item Improved optimization of skip connections and attention units within an autoencoder framework.
  \item Effective multi-visual fusion, enabling better detection of anomalies with multiple camera modalities.
\end{itemize}

The paper's contribution of developing Temporal Shift to the field of video anomaly detection is a significant advancement that has the potential to be applied beyond fall detection to other anomaly detection applications.

\section{Literature Review}

Video anomaly detection has a wide range of applications and methods outside of fall detection. Following the success of deep learning in image classification \cite{krizhevsky2017imagenet, he2016deep}, many researchers have started focusing on the anomaly detection problem in videos by applying different types of deep learning methods, such as 3D Convolutional Autoencoders (CAEs) \cite{hu2022video, mishra2023privacy},  Long Short-Term Memory Networks (LSTMs) \cite{abbas2022comprehensive} and Temporal Convolutional Networks \cite{abedi2023detecting}. These approaches typically involve training the model on normal data and then detecting deviations in the reconstruction error or output to identify anomalies.

\subsection{Video Anomaly Detection for Falls}

Traditional handcrafted features for fall detection rely on changes in bounding box proportions or joint/skeletal positional information to extract features characterizing the body's geometry, motion, and position \cite{ramachandran2020survey, gutierrez2021comprehensive}. However, certain limitations, such as occlusions or camera positioning within homes \cite{Baldewijns2016}, can hinder the robustness and accuracy of these methods in fall detection. Deep learning methods that can extract a broader range of features might offer improved performance.

Nogas et al. \cite{Nogas2020} introduced a 3D Convolutional Autoencoder (3DCAE) for fall detection within an anomaly detection framework. The reconstruction error from the 3DCAE was used directly to detect falls. This method was found to outperform convolutional LSTMs \cite{nogas2018fall}. Khan et al. \cite{khan2021spatio} extended this work by incorporating the autoencoder into an adversarial learning system, demonstrating similar performance using thermal and depth cameras. Mehta et al. \cite{mehta2021motion} added a secondary stream that encodes optical flow images, placing more emphasis on the temporal component. Moreover, a region-aware mechanism was used to isolate the image regions where the body is located, creating a region of interest.

Ronneberger et al. proposed a U-Net architecture for biomedical image segmentation \cite{ronneberger2015u}. This method has been extrapolated for use with 3D images through a V-Net \cite{milletari2016v}. Recently, a U-Net has also been used in a video anomaly detection framework \cite{kim2022video}. Oktay et al. \cite{oktay2018attention} introduced an Attention U-Net to improve upon existing U-Net structures. Attention Gates (AGs) enable a model to focus on specific structures within the image without additional supervision. Attention mechanisms form the backbone of state-of-the-art models such as transformers and have been widely used in predictive tasks, including natural language processing \cite{vaswani2017attention, devlin2018bert}. Since skip connections in a U-Net structure convey abundant direct spatial and temporal information, AGs provide a way to isolate important aspects of the image. However, in the context of reconstruction error, all parts of an image are equally important for reducing error.

Applying certain techniques proven effective in other deep learning computer vision tasks, such as U-Net structures, Residual connections, or Attention Gates, to video anomaly detection can be challenging. These techniques, which modify the information flow through the network, were developed primarily for classification or segmentation tasks. When reconstruction error is used as the loss function, it prompts these network structures to propagate as much information as possible, bypassing much of the learning process. As such, a modified loss function, such as proposed by Liu et al. \cite{liu2018future}, can help incorporate these methods within a video anomaly detection framework.

\subsection{Future Frame Prediction}

Several researchers have used the prediction of a single or multiple future frames (or images in a video sequence) for video anomaly detection tasks. Liu et al. \cite{liu2018future} used a stack of four frames to predict the fifth frame. The adversarial loss is found by discriminating between the predicted optical flow frames and the true frame. This means that only the predicted frame is used to find the loss, and none of the reconstructed frames. However, this still allows for optimization of the U-Net skip connections introduced in the Generative Adversarial Network (GAN). The authors normalized all of the reconstruction errors between 0 and 1. This introduces temporal leakage, as future values were used to normalize past ones. The approach of using four frames and then predicting the following frame was also used by Nguyen et al. \cite{nguyen2019anomaly} in a two-stream network. Jamadandi et al., \cite{jamadandi2018predgan} use a GAN to take a series of four frames and predicts the subsequent four (future) frames. Wu et al. \cite{wu2019deep} also used four frames to predict the fifth in their latent space restriction approach. Tang et al., \cite{tang2020integrating} introduced a method that also seeks to use reconstruction in addition to the prediction error of frames to detect anomalies. Their approach also used optical flow images and the four-part loss function introduced by Liu et al. \cite{liu2018future}. However, they use two U-Net autoencoders in series. The first autoencoder takes an input of four frames \( I_{t} \), \( I_{t+1} \), \( I_{t+2} \), \( I_{t+3} \) and seeks to predict the $5^{th}$ frame, or \( I_{m} \). The second autoencoder then takes this predicted and tries to reconstruct the ground truth or \( I_{t+4} \). However, this method only tries to predict a single frame. We expand on this work by introducing temporal shift, a multi-objective loss function that uses both the reconstruction and prediction of frames. This is accomplished by predicting an overlapping window of frames, of which a portion of frames are reconstructed and the rest is predicted.

\section{Methods} 

The methods section will cover three primary aspects: the dataset utilized, the proposed loss function, and the evaluation methods employed for the loss function. The evaluation will involve three main models. The first model is a baseline 3D convolutional autoencoder (3DCAE). 
Next, the 3DCAE architecture is adapted into an Attention U-Net. Lastly, the 3DCAE is modified into a multi-modal network to handle multiple camera modalities for fall detection. The objective is to assess how the loss function, Temporal Shift, enhances the optimization of different network types within an anomaly detection framework for the purpose of fall detection with single and multiple camera modalities.

\subsection{Dataset}

The Multi Visual Modality Fall Detection Dataset (MUVIM) was used for anomaly detection of falls \cite{denkovski2022multi}. It contains (6) vision-based sensors of different modalities including thermal, depth, infra-red (IR) and RGB cameras (See Figure \ref{fig:activity_2} for example images). Pre-processing steps included converting images to grey scale, re-sizing to 64 by 64, normalizing image values and reducing frame rate to 8 frames per second. In the case of the thermal camera, the frame rate was significantly lower. As such, frames were duplicated as needed in order interpolate the dataset to 8 frames per second. The dataset contains videos of older adults performing activities of daily living (ADL) activities and is used as the normal dataset for training. Videos containing both normal activities and falls are performed by younger adults in the 
test set. The dataset consisted of 200,000 frames, with 160,000 frames used for testing and 40,000 frames used for training. The dataset collected contains both older adults (with normal activities and no falls) and younger adults (with both normal activities and falls). Only video's without falls were used for training to define the normal class, thus only older adults' data was used for training and younger adults' data for testing. This division ensures the model differentiates between fall and non-fall events without being influenced by age-related differences, preventing model bias. In an anomaly detection approach, the data for “anomalous” class (fall in this case) is not available; therefore, no validation set is explicitly created to avoid information leakage during testing.

The combination of multiple modalities was of interest as its been observed that the addition of visual modalities such as RGB and depth can improve performance \cite{song2015sun, eitel2015multimodal}. Modalities selected for combination were limited to those captured by the same camera system - such as the Orbbecc Depth and Infra-red images. This allows for the highest degree of similarities across modalities. 

\begin{figure}[!htbp]
\floatconts
{fig:activity_2}
{\caption{Start of the fall as indicated by manually produced labels for each camera. (a) Center FLIR Thermal, (b) Hikvision IP, (c) Orbbec Depth, (d) Orbbec IR, (e) - Stereolabs ZED Depth, and (f) Stereolabs ZED  RGB }}
    {%
        \subfigure{%
        \label{fig:pic1}
        \includegraphics[width=0.25\textwidth]{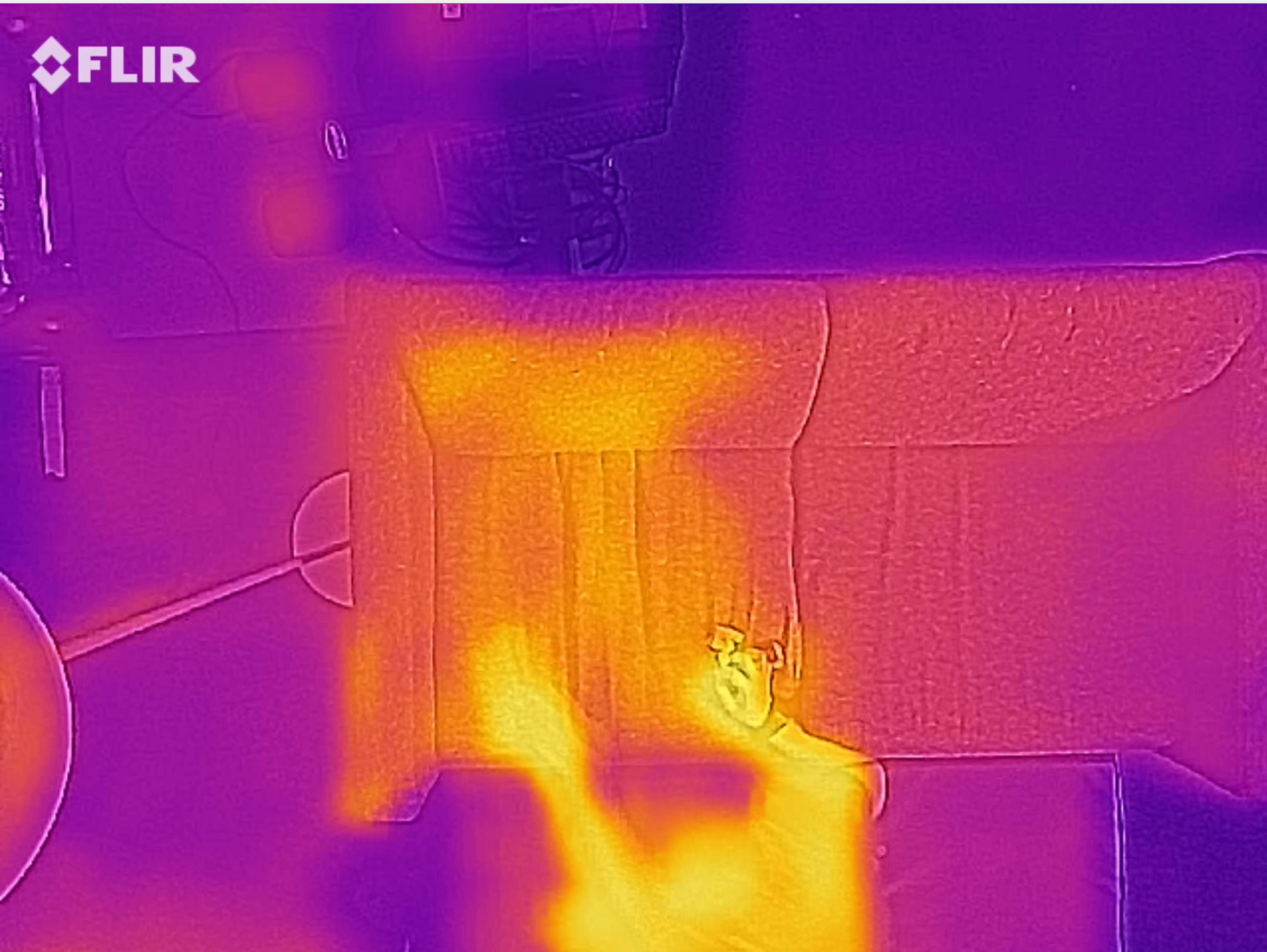}
        }\qquad 
        \subfigure{%
        \label{fig:pic2}
        \includegraphics[width=0.25\textwidth]{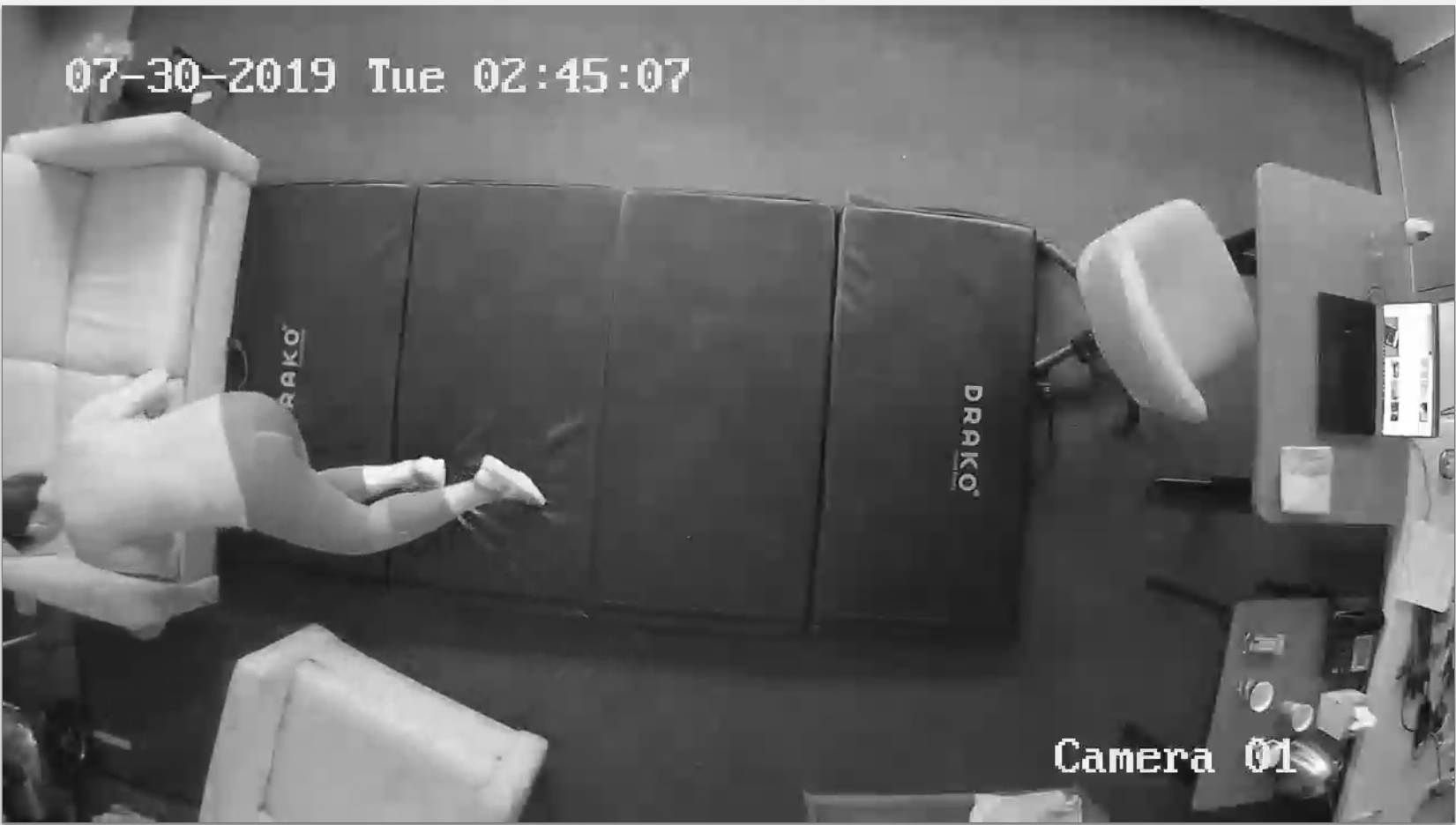}
        }\qquad
        \subfigure{%
        \label{fig:pic3}
        \includegraphics[width=0.25\textwidth]{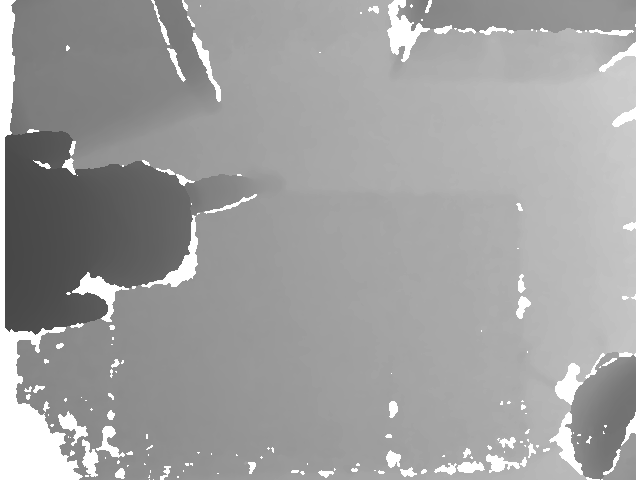}
        } 
        
        \subfigure{%
        \label{fig:pic4}
        \includegraphics[width=0.25\textwidth]{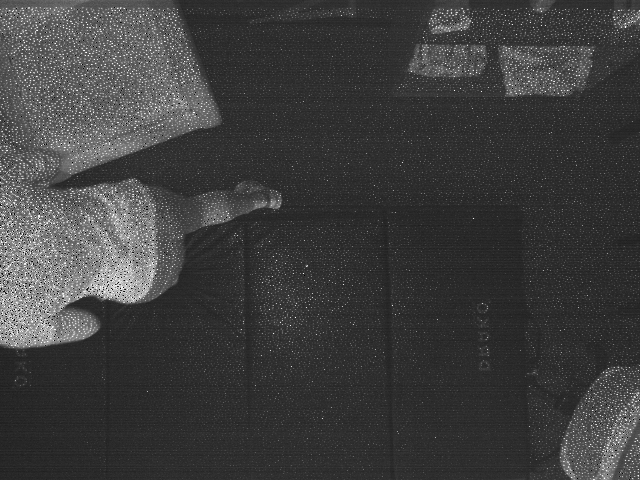}
        }\qquad
        \subfigure{%
        \label{fig:pic5}
        \includegraphics[width=0.25\textwidth]{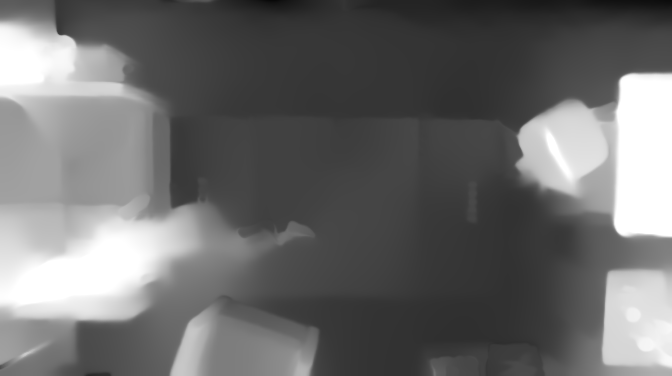}
        }\qquad 
        \subfigure{%
        \label{fig:pic6}
        \includegraphics[width=0.25\textwidth]{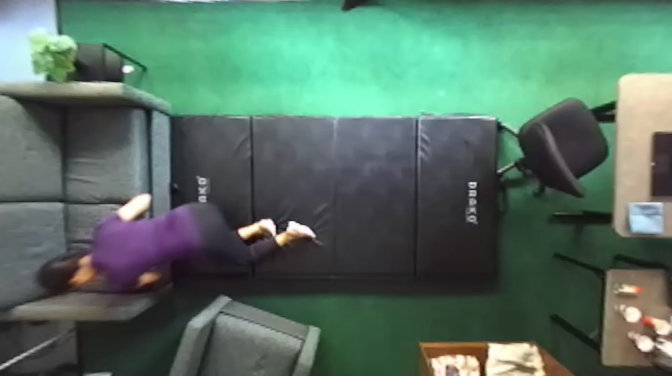}
        }
        
    }
\end{figure}

\subsection{Temporal Shift Loss}

\begin{figure*}[htpb]
    \centering
    \includegraphics[width=0.85\textwidth]{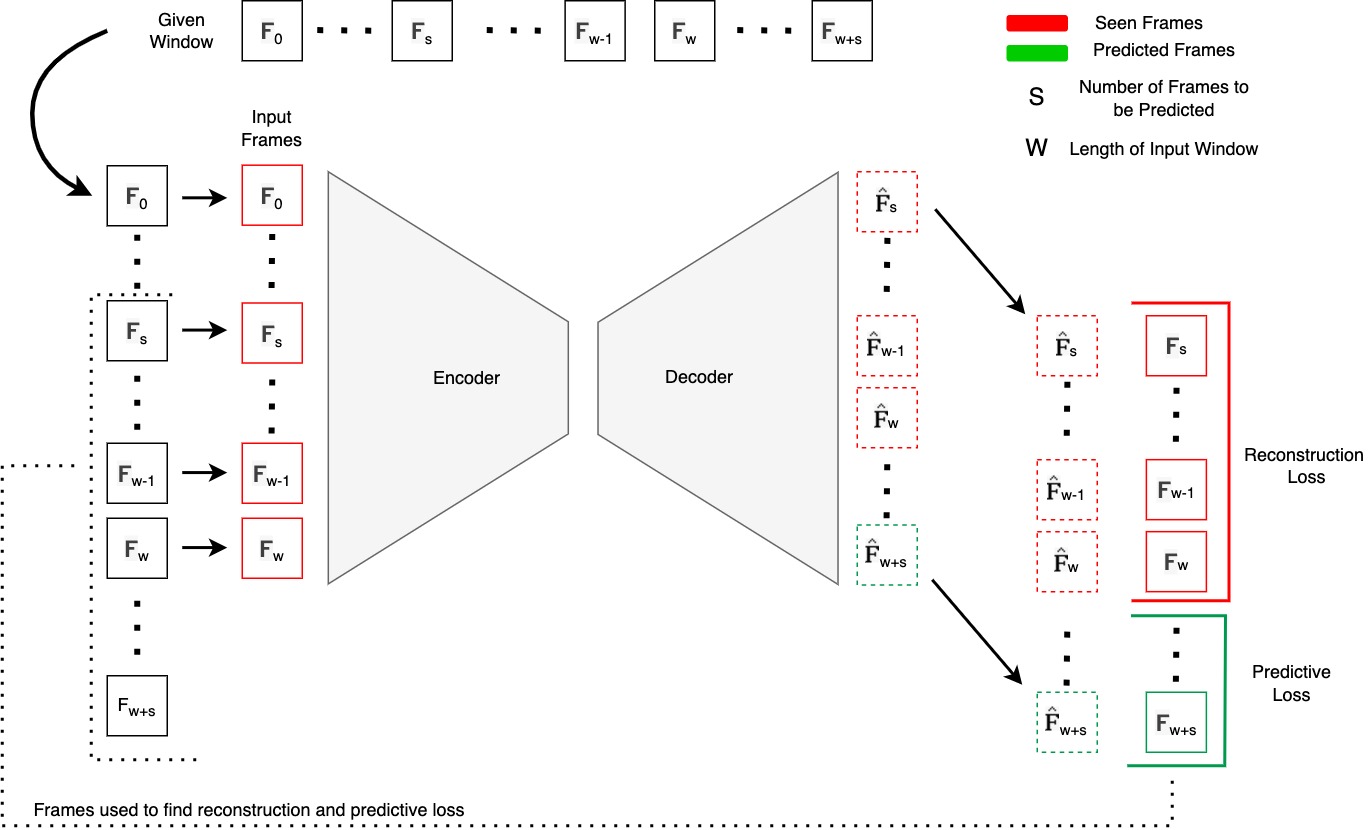}
    \caption{The six input frames (F) and six output frames (P) for a window length of eight, and a temporal shift of two.}
    \label{fig:temporal_pred} 
\end{figure*}

In the context of an autoencoder, temporal shift is a multi-objective problem comprised of reconstruction loss and prediction loss. The first part, reconstructive loss, is found by comparing the reconstruction error of frames that were input into the autoencoder with reconstructions of those same frames. Prediction loss is found by the autoencoder generating frames that were 
not input into the autoencoder and are instead only used to find the error. See Figure \ref{fig:temporal_pred} for details.

\begin{figure}[!htpb]
    \centering
    \includegraphics[scale=0.5]{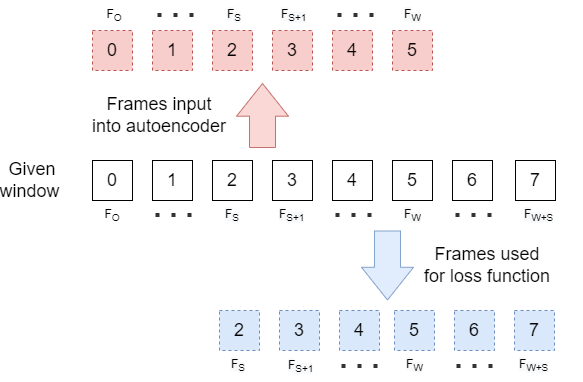}
    \caption{Example of frame selection for a window of 8 frames, and an input window of \( W = 6 \) and a temporal shift of \( S = 2 \)}
    \label{fig:frameselection} 
\end{figure}

For a given window of length \( L \), it is divided into two overlapping sub-windows. The first sub-window is comprised of frames \( 0 \) to \( W \). The second sub-window is temporally shifted by an amount equal to \( S \). This means the second sub-window contains frames \( S \) to \( W + S \), where \( L = W + S \). In essence, we are generating a window of frames (with length \( W \)) that has been shifted by a temporal offset of \( S \). For example, if we want an input window of 6 frames  (\( W = 6 \) ) and a temporal shift \( S \) of 2, we will need a total window size of \( W + S \) or 8. Then we input the first sub-window (frames 0 to 5) into the autoencoder and use the second sub-window to find reconstruction error (frames 2 to 7). This will result in having 4 frames that are reconstructed (2 to 5) and 2 that are predicted (6 and 7). See Figure \ref{fig:frameselection}

\subsubsection{Temporal Shift - Hyper Parameter Tuning}

Hyper parameter tuning on an anomaly detection dataset is very challenging. This is because in an anomaly detection framework, it is assumed that one does not have the anomalous class data. As such, the use of a validation set cannot be used. However, tuning the hyperparameter on the test set is problematic due to the model to adapting to fit (or over-fit) test set and also due to leakage of information. 
Recognizing this, several methods were tried to 
identify the best performance on the test set. 

There are two main hyper-parameters that need to be set. The first is the length of the window. It was found that eight frames per window worked best. However, this may be dependent on video frame rate and the specifics of anomaly (fall in our case). The second parameter is the length of time shift and thus the number of frames to be predicted. Results for various window lengths are shown in Table \ref{table:windowlengthcomparison} in the results section. To compare to related work by \cite{liu2018future, nguyen2019anomaly}, the window length was set to five and the temporal shift to one. This meant that four frames would be used to predict one frame. Additionally, window lengths 6, 8 and 10 were explored with various temporal shifts of 1 through 4. Even numbers were only used to eliminate rounding scenarios. Experimentally, it was found that the best performance was achieved with an input window of 6 and a temporal shift of 2, so these were used for further experiments. This means a total window length of 8 was used.

\subsection{Modeling}

Three main variations of modeling approaches were employed in this study. The first two variations focus on a single modality at a time. These include the baseline 3D convolutional autoencoder (3DCAE) and the Attention U-Net networks. The 3DCAE serves as a baseline model structure, allowing us to evaluate the impact of the Temporal Shift Loss on a network without skip connections. The Attention U-Net, on the other hand, incorporates attention mechanisms and skip connections in order to evaluate if temporal shift loss can effectively optimize the networks.
The third variation involves a multi-modal network that examines whether combining two video modalities can enhance performance compared to using a single modality. This approach simultaneously incorporates two modalities and combines them in the latent space. While the combination of multiple modalities, such as RGB and Depth, has been explored in classification tasks, it remains to be seen whether incorporating multiple modalities can provide a better definition of reconstruction error-based loss in the context of anomaly detection.



\subsubsection{3DCAE}


The 
3DCAE has demonstrated superior performance in fall detection within an anomaly detection framework \cite{Nogas2020}. It was found to outperform other methods, including convolutional LSTMs \cite{nogas2018fall}. The 3DCAE has also shown to be adaptable to other frameworks, such as adversarial learning \cite{khan2021spatio}. 
We chose to use a 3DCAE as our reference baseline. Compared to other deep learning methods, this approach offers high performance while maintaining relative simplicity. Its adaptability enables the integration of other network structures, such as skip connections or attention gates relatively easily. This adaptability facilitates the assessment of the impact of these network structures when coupled with and without Temporal Shift. The MUVIM dataset was introduced by Denkovski et al. (2022), who conducted extensive experiments on window sizes and network hyper-parameters. Given our utilization of the same dataset and baseline autoencoder model, we adopted their model parameters.


The models are only trained on normal data. Consequently, when a fall event occurs, the reconstruction in expected to be high in majority of the cases. During the testing phase, the reconstruction error is calculated for a given window. However, as a frame may appear in multiple windows, the frame's reconstruction error is determined by averaging the reconstruction errors from all windows it appears in. The reconstruction error from all videos are then concatenated into a large vector. Now to determine when a fall has happened, a threshold for the reconstruction error would have to be established. However, this creates an inherent trade-off between sensitivity and specificity (i.e. can choose a lower false positive rate but higher true positive rate). To circumvent this trade-off, the performance evaluation utilizes the area under the curve metrics, specifically the receiver operating characteristic curve and the precision-recall curve. By employing these metrics, the performance assessment encompasses all possible thresholds without the necessity of setting a specific threshold.

 \begin{figure}[]
    \centering
    \includegraphics[scale=0.35]{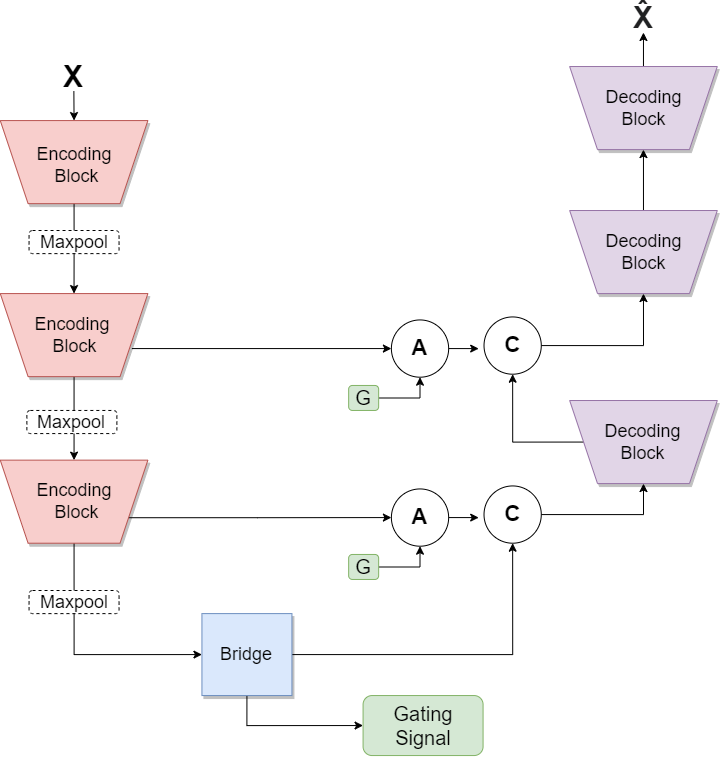}
    \caption{Attention U-Net Structure used. Circle A; The modified attention gate mechanism. Gating signals from the lowest (coarsest) dimensionality were up-scaled in order to be applied to various feature maps.}
    \label{fig:attentionunet} 
\end{figure}

\subsubsection{Attention U-Net}


The overall structure for the Attention U-Net is shown in Figure \ref{fig:attentionunet}. As outlined earlier, skip connections and attention structures usually don't perform well in networks that focus on reconstruction. In such cases, learning can be bypassed and all information is important to minimizing reconstruction loss. A few key modifications to the original U-Net structure were implemented to help prevent direct propagation of information. Firstly, the last U-Net skip connection was removed, preventing access to information from the lowest levels. Second, the sigmoid layer used in the attention gate was replaced with a sequential softmax activation across each dimension. This was done as it causes the weighting of the features to become sparser as they must sum to 1. This would prevent a heavy weighting on all features, which may occur if a sigmoid activation layer was used in a reconstruction framework. The encoder consisted of three encoding blocks connected to a bridge. The decoder was comprised of three decoding blocks; two of which have attention gated skip connections directly connecting them to encoding blocks. As done in the original Attention U-Net paper \cite{schlemper2019attention}, the gating signal was taken from the coarsest layer and up-scaled to fit various feature maps.


\subsubsection{Multi-Modal Network}

 \begin{figure}[!htbp]
    \centering
    \includegraphics[scale=0.25]{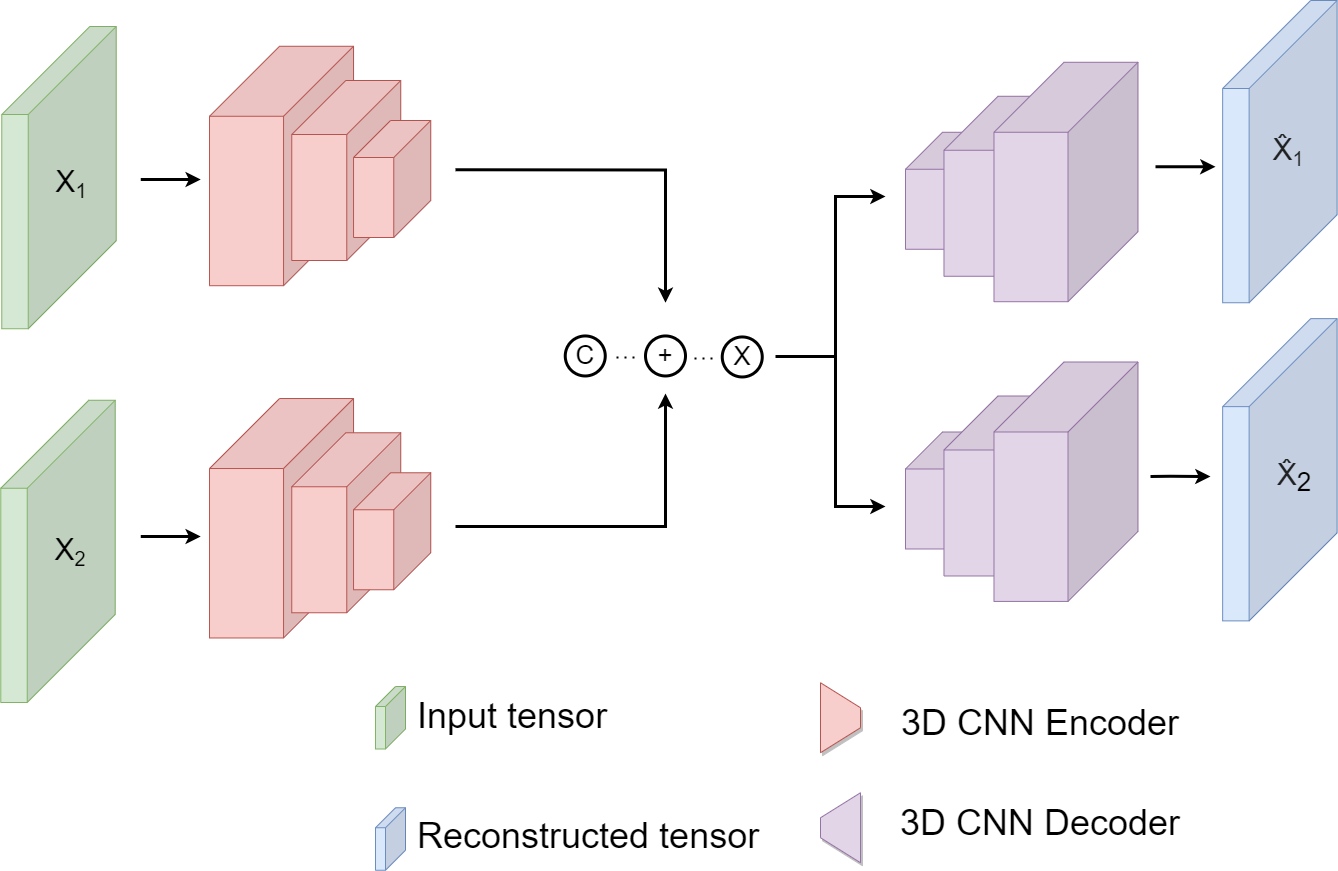}
    \caption{Multi-modal network framework used. Fusion occurred by one of three operations (multiplication, addition, or concatenation) at the bottleneck layer.}
    \label{fig:multimodalnetwork} 
\end{figure}

The multi-modal framework is outlined in Figure \ref{fig:multimodalnetwork}. It is comprised of individual encoders and decoders. The baseline 3DCAE network structure is used. The modalities are combined in the latent space through one of three operations, addition, multiplication, or concatenation. The addition or multiplication of feature maps amplifies the contribution of shared features that are heavily weighted, which increases their importance in the final output. This can improve the performance of the model by allowing it to focus on the most important features and make more accurate predictions. However, with an anomaly detection framework sharing of features may be difficult as decoding is focused on reconstructing individual modalities. In addition the low-level features such as shapes, edges, are shared between the modalities with only high-level features being different.
Models are only trained on normal or ADL data consisting of older adults. Test data contains a mixture of ADL data and falls from younger adults. As a single frame appears multiple times across overlapping windows, the mean reconstruction error for a frame is used.

\section{Results}


\begin{table*}[!htbp]
    \sisetup{group-minimum-digits = 4}
    \centering
    \caption{Results achieved by various temporal windows with the baseline 3D CNN model. W represents the length of the input window and S represents the amount of temporal shift. The total window length is the sum of these two.}
    \label{table:windowlengthcomparison}
    \resizebox{\linewidth}{!}{%
    \begin{tabular}{ccccccccc}
        \toprule[1pt]\midrule[0.3pt]
        \thead{\multicolumn{1}{l}{W}} & \multicolumn{1}{l}{S} & AUC & Hikvision IR & Orbbec IR & Orbbec Depth & ZED Depth & ZED RGB & Thermal \\ 
        \midrule
        \multirow{2}{*}{8}               & \multirow{2}{*}{0}            & ROC & 0.904 & \textbf{0.905} & 0.820 & 0.795 & 0.703 & \textbf{0.831} \\ 
                                         &                               & PR  & 0.074 & 0.068 & 0.031 & 0.028 & 0.018 & 0.061 \\[0.1cm] 
        
        \multirow{2}{*}{6}               & \multirow{2}{*}{2}            & ROC & \textbf{0.912} & 0.902 & \textbf{0.824} & \textbf{0.834} & \textbf{0.729} & 0.814 \\
                                         &                               & PR  & 0.075 & \textbf{0.076} & 0.033 & \textbf{0.034} & \textbf{0.021} & \textbf{0.062} \\[0.1cm]
        
        \multirow{2}{*}{4}               & \multirow{2}{*}{4}            & ROC & \textbf{0.912} & 0.900 & 0.775 & 0.776 & 0.684 & 0.806 \\
                                         &                               & PR  & \textbf{0.091} & 0.069 & 0.026 & 0.029 & 0.021 & 0.057 \\[0.1cm]  
                                         
        \multirow{2}{*}{4}               & \multirow{2}{*}{1}            & ROC & 0.826 & 0.872 & 0.744 & 0.649 & 0.612 & 0.643 \\ 
                                         &                               & PR  & 0.058 & 0.053 & \textbf{0.044} & 0.015 & 0.016 & 0.014\\
                                         \midrule[0.3pt]\bottomrule[1pt]
    \end{tabular}
    }
\end{table*}

\begin{table*}[!htbp]
  \sisetup{group-minimum-digits = 4}
  \centering
  \caption{AUC ROC and PR values for various loss functions. No Skill shows the performance achieved by a random chance classifier. All results are found with a window of eight frames and a temporal shift of two. Bold values indicate the highest results for that modality. Italic script indicates the loss function used for training and testing.}
  \label{tab:aucpr}
  \resizebox{\linewidth}{!}{%
  \begin{tabular}{lccccccc} 
    \toprule[1pt]\midrule[0.3pt]
    
    \thead{Visual Modality\\[0.1cm]} & \thead{AUC} & \thead{No Skill} & \thead{3D CNN\\[0.1cm]  \textit{Reconstruction}} & \thead{3D CNN\\[0.1cm]  \textit{Prediction}} & {\thead{3D CNN\\[0.1cm]  \textit{Temporal Shift}}} & \thead{Attention U-Net\\[0.1cm]  \textit{Reconstruction}} & \thead{Attention U-Net\\[0.1cm]  \textit{Temporal Shift}}  \\

    \midrule

     Orbbec IR & ROC & 0.5 & 0.905 & 0.866 & 0.902 & 0.816 & \textbf{0.920} \\
         & PR & 0.010 & 0.068 & 0.048 & 0.076 & 0.074 & \textbf{0.105} \\[0.2cm]
         
    Orbbec Depth & ROC & 0.5 & 0.820 & 0.659 & 0.824 & 0.700 & \textbf{0.829} \\
         & PR & 0.010 & 0.031 & 0.014 & 0.033 & 0.033 & \textbf{0.039} \\[0.2cm]
         
    ZED Depth & ROC & 0.5 & 0.795 & 0.590 & \textbf{0.834} & 0.616 & 0.724 \\
         & PR & 0.017 & 0.028 & 0.012 & \textbf{0.034} & 0.012 & 0.029 \\[0.2cm]
         
    ZED RGB & ROC & 0.5 & 0.703 & 0.651 & \textbf{0.729} & 0.612 & 0.726 \\
         & PR & 0.017 & 0.018 & 0.014 & 0.021 & 0.017 & \textbf{0.022} \\[0.2cm]
         
    Thermal & ROC & 0.5 & \textbf{0.831} & 0.769 & 0.814 & 0.612 & 0.786 \\
         & PR & 0.009 & 0.061 & 0.047 & \textbf{0.062} & 0.045 & 0.058 \\[0.2cm]
         
    Hikvision IR & ROC & 0.5 & 0.904 & 0.784 & \textbf{0.912} & 0.895 & \textbf{0.912} \\
         & PR & 0.010 & 0.068 & 0.048 & 0.076 & 0.074 & \textbf{0.105} \\[0.2cm]

    \midrule[0.3pt]\bottomrule[1pt]

  \end{tabular}
    }
\end{table*}

Area under the curve of receiver operating characteristics (AUC ROC) and precision recall curve (AUC PR) are reported per modality in Table \ref{tab:aucpr}. The results of hybrid fusion are reported in Table \ref{tab:multimodal}. 

Table \ref{table:windowlengthcomparison} shows the comparison of results with various choices of W and S. It can be observed that W=6 and S=2 works best for most of the camera modalities for detecting falls. We keep this configuration for the rest of the experiments. 
From Table \ref{tab:aucpr}, we observe that temporal shift or the use of both reconstruction and prediction loss can improve fall detection results, with the highest AUC ROC and AUC PR scores achieved by models that incorporate both loss functions in temporal shift.  The highest AUC ROC was achieved with the Orbbec IR camera with Attention U-Net with an AUC ROC of 0.920 and also achieved the highest AUC PR of 0.105. This has improved on baseline results by 0.02 AUC ROC and 0.03 AUC PR. Larger performance increases are seen in a multi-modal framework as seen in Table \ref{tab:multimodal}. Two of the three combinations of modality saw large performance increases, with multiplication bottleneck fusion performing the best. The performance increased from 0.893 AUC ROC to 0.929 AUC ROC for the combination of both Orbbec Infrared and Depth cameras. The combination of Hikvision IR and ZED Depth cameras saw a performance improvement of 0.855 to 0.901 AUC ROC. AUC PR also increased with the combination of Orbbec cameras, achieving the highest AUC PR score of 0.126.

\begin{table*}[!htbp]
\sisetup{group-minimum-digits = 4}

\centering
\caption{Performance of various multi-modal fusion methods. Columns indicate the modalities/cameras used, and each models respective performance with different loss functions. Best performance per input modality combination is bolded.}
\label{tab:multimodal}
\resizebox{\linewidth}{!}{%
    \begin{tabular}{llcccccc} 
        \toprule[1pt]\midrule[0.3pt]
        \multicolumn{1}{c}{Model} & \multicolumn{1}{c}{AUC} & \multicolumn{2}{c}{\begin{tabular}[c]{@{}c@{}}Orbbec IR \\+ Orbbec Depth\end{tabular}} & \multicolumn{2}{c}{\begin{tabular}[c]{@{}c@{}}ZED Depth \\+ ZED RGB\end{tabular}} & \multicolumn{2}{c}{\begin{tabular}[c]{@{}c@{}}Hikvision IR~\\+ ~ZED Depth\end{tabular}} \\
        \multicolumn{1}{c}{} & \multicolumn{1}{c}{} & \textit{Reconstruction} & \textit{Temporal Shift} & \textit{Reconstruction} & \textit{Temporal Shift} & \textit{Reconstruction} & \textit{Temporal Shift} \\ 
        
        \midrule
        
        \multirow{2}{*}{Hybrid Concat} & ROC & 0.886 & 0.924 & \textbf{0.787} & 0.716 & 0.752 & 0.890 \\
         & PR & 0.051 & 0.111 & \textbf{0.059} & 0.018 & 0.021 & 0.068 \\[0.1cm]
        \multirow{2}{*}{Hybrid Add} & ROC & 0.858 & 0.927 & 0.760 & 0.733 & 0.833 & 0.896 \\
         & PR & 0.039 & 0.125 & 0.019 & 0.021 & 0.046 & 0.092 \\[0.1cm]
        \multirow{2}{*}{Hybrid Multi} & ROC & 0.893 & \textbf{0.929} & 0.758 & 0.746 & 0.855 & \textbf{0.901} \\
         & PR & 0.080 & \textbf{0.126} & 0.020 & 0.022 & 0.055 & \textbf{0.103} \\
    
         \midrule[0.3pt]\bottomrule[1pt]
    
    \end{tabular}
    }
\end{table*}

\subsection{Discussion}

The implementation of temporal shift has been shown to improve the performance of video anomaly detection for falls, specifically for 3DCAE, Attention U-Net, and multi-modal models across different datasets. In particular, Attention U-Nets showed a large increase of almost 0.20 AUC ROC, indicating that temporal shift allows for more effective optimization of techniques used in classification, such as skip connections and attention gates in an anomaly detection framework. The thermal camera may not have improved as much as others, because it had duplicated frames in order to increase the effective frame rate. 

Multi-modal models also showed an improvement in performance when temporal shift was applied. Without this method, these models tend to perform worse than single modalities. This may have been due to difficulties in leveraging shared features for reconstruction in multi-modal autoencoders, but the additional information provided by the different modalities may have been beneficial in prediction. Only two modalities were combined at a time due to computational limitations and practical considerations, and these modalities were captured by the same camera to account for frame rate and manual labeling differences. Although improved performance was seen when Depth and IR camera modalities were combined, a decrease in performance was seen with Depth and RGB. This may have been due to the fact the RGB camera was the worst performing modality and may have limited performance benefits.

The overall performance of the model, measured by the AUC ROC metric, is high. The baseline AUC PR performance is dependent on the distribution of labels within the dataset. Despite this, the best performing models demonstrated a significant improvement over random chance, with an approximate ten-fold increase in AUC PR. However, the absolute performance of the model remains low. This may be due to limitations in the dataset for anomaly detection, specifically the lack of diversity in the normal set of data, which indicates moderately higher false positive rates. 

Experiments on different window lengths and frames consistently showed that predicting multiple frames outperformed predicting a single frame. The optimal results for fall detection were found with 8 frames (1 second window) and a temporal shift of 2 frames (0.25 seconds). However, if the window size is too long and the temporal length of the prediction is extended too far into the future, the task may become too difficult and less beneficial. The length of input windows and temporal shift used may depend on the type of anomaly and frame rate of the dataset.

\section{Conclusions and Future Work}

The implementation of temporal shift has been shown to greatly improve the performance of video anomaly detection for falls. Notably, the attention U-Net gave an AUC ROC increase of 20\%, indicating better optimization of classification modeling techniques within anomaly detection framework. Both the multi-modal model and baseline 3DCAE model benefited from the implementation of temporal shift, likely because predicting multiple frames requires a better understanding of temporal feature changes over time. The temporal shift approach may work better with anomalies that are sudden in nature, e.g., falls. 

In future, we will evaluate the usefulness of slowly evolving anomalies with our approach. The performance of temporal shift could be improved using different network architectures and attention mechanisms.  Additionally, exploring different weightings for the reconstruction loss and prediction loss could provide valuable insights. The temporal shift approach can be applied to many applications, including violence detection in crowded scenes, detecting rare diseases in 3D medical images and responsive behaviors in people with dementia using CCTV footage. 

Video anomaly detection frameworks offer can help with class imbalance problems like fall detection, but they have limitations. Anything not in the training set is considered anomalous, leading to high false positives and low AUC PR. This study used a semi-naturalistic dataset, which may not fully represent real-world falls. A real world dataset may contain a wider range of activities allowing more efficient model training. Future work will develop privacy-protecting models and establish thresholds for fall detection and alarm triggering.

Temporal shift allowed for the effective training of network structures that use skip connections without a classification based loss function. We hope that this method may be beneficial in many other tasks. 

\section{Acknowledgements}
The research work presented in this paper was supported by Natural Sciences and Engineering Research Council of Canada.

\bibliography{references}

\begin{thebibliography}{39}
\providecommand{\natexlab}[1]{#1}
\providecommand{\url}[1]{\texttt{#1}}
\expandafter\ifx\csname urlstyle\endcsname\relax
  \providecommand{\doi}[1]{doi: #1}\else
  \providecommand{\doi}{doi: \begingroup \urlstyle{rm}\Url}\fi

\bibitem[Abbas and Al-Ani(2022)]{abbas2022comprehensive}
Zainab~K Abbas and Ayad~A Al-Ani.
\newblock A comprehensive review for video anomaly detection on videos.
\newblock In \emph{2022 International Conference on Computer Science and
  Software Engineering (CSASE)}, pages 1--1. IEEE, 2022.

\bibitem[Abedi and Khan(2023)]{abedi2023detecting}
Ali Abedi and Shehroz~S. Khan.
\newblock Detecting disengagement in virtual learning as an anomaly using
  temporal convolutional network autoencoder.
\newblock \emph{Signal, Image and Video Processing}, pages 1--9, 2023.

\bibitem[Baldewijns et~al.(2016)Baldewijns, Debard, Mertes, Vanrumste, and
  Croonenborghs]{Baldewijns2016}
Greet Baldewijns, Glen Debard, Gert Mertes, Bart Vanrumste, and Tom
  Croonenborghs.
\newblock {Bridging the gap between real-life data and simulated data by
  providing a highly realistic Fall dataset for evaluating camera-based fall
  detection algorithms}.
\newblock \emph{Healthcare Technology Letters}, 3\penalty0 (1):\penalty0 6--11,
  mar 2016.
\newblock ISSN 20533713.
\newblock \doi{10.1049/htl.2015.0047}.
\newblock URL \url{https://onlinelibrary.wiley.com/doi/10.1049/htl.2015.0047}.

\bibitem[Charfi et~al.(2012)Charfi, Miteran, Dubois, Atri, and
  Tourki]{charfi2012definition}
Imen Charfi, Johel Miteran, Julien Dubois, Mohamed Atri, and Rached Tourki.
\newblock Definition and performance evaluation of a robust svm based fall
  detection solution.
\newblock In \emph{2012 eighth international conference on signal image
  technology and internet based systems}, pages 218--224. IEEE, 2012.

\bibitem[Debard et~al.(2012)Debard, Karsmakers, Deschodt, Vlaeyen, Dejaeger,
  Milisen, Goedem{\'e}, Vanrumste, and Tuytelaars]{debard2012camera}
Glen Debard, Peter Karsmakers, Mieke Deschodt, Ellen Vlaeyen, Eddy Dejaeger,
  Koen Milisen, Toon Goedem{\'e}, Bart Vanrumste, and Tinne Tuytelaars.
\newblock Camera-based fall detection on real world data.
\newblock In \emph{Outdoor and large-scale real-world scene analysis}, pages
  356--375. Springer, 2012.

\bibitem[Denkovski et~al.(2022)Denkovski, Khan, Malamis, Moon, Ye, and
  Mihailidis]{denkovski2022multi}
Stefan Denkovski, Shehroz~S Khan, Brandon Malamis, Sae~Young Moon, Bing Ye, and
  Alex Mihailidis.
\newblock Multi visual modality fall detection dataset.
\newblock \emph{IEEE Access}, 10:\penalty0 106422--106435, 2022.

\bibitem[Devlin et~al.(2018)Devlin, Chang, Lee, and Toutanova]{devlin2018bert}
Jacob Devlin, Ming-Wei Chang, Kenton Lee, and Kristina Toutanova.
\newblock Bert: Pre-training of deep bidirectional transformers for language
  understanding.
\newblock \emph{arXiv preprint arXiv:1810.04805}, 2018.

\bibitem[Eitel et~al.(2015)Eitel, Springenberg, Spinello, Riedmiller, and
  Burgard]{eitel2015multimodal}
Andreas Eitel, Jost~Tobias Springenberg, Luciano Spinello, Martin Riedmiller,
  and Wolfram Burgard.
\newblock Multimodal deep learning for robust rgb-d object recognition.
\newblock In \emph{2015 IEEE/RSJ International Conference on Intelligent Robots
  and Systems (IROS)}, pages 681--687. IEEE, 2015.

\bibitem[Fuster(2017)]{fuster2017changing}
Valentin Fuster.
\newblock Changing demographics: a new approach to global health care due to
  the aging population, 2017.

\bibitem[Guti{\'e}rrez et~al.(2021)Guti{\'e}rrez, Rodr{\'\i}guez, and
  Martin]{gutierrez2021comprehensive}
Jes{\'u}s Guti{\'e}rrez, V{\'\i}ctor Rodr{\'\i}guez, and Sergio Martin.
\newblock Comprehensive review of vision-based fall detection systems.
\newblock \emph{Sensors}, 21\penalty0 (3):\penalty0 947, 2021.

\bibitem[He et~al.(2016)He, Zhang, Ren, and Sun]{he2016deep}
Kaiming He, Xiangyu Zhang, Shaoqing Ren, and Jian Sun.
\newblock Deep residual learning for image recognition.
\newblock In \emph{Proceedings of the IEEE conference on computer vision and
  pattern recognition}, pages 770--778, 2016.

\bibitem[Hu et~al.(2022)Hu, Lian, Zhang, Gao, Jiang, and Chen]{hu2022video}
Xing Hu, Jing Lian, Dawei Zhang, Xiumin Gao, Linhua Jiang, and Wenmin Chen.
\newblock Video anomaly detection based on 3d convolutional auto-encoder.
\newblock \emph{Signal, Image and Video Processing}, pages 1--9, 2022.

\bibitem[Igual et~al.(2013)Igual, Medrano, and Plaza]{Igual2013}
Raul Igual, Carlos Medrano, and Inmaculada Plaza.
\newblock {Challenges, issues and trends in fall detection systems}.
\newblock Technical Report~1, 2013.
\newblock URL
  \url{http://www.biomedical-engineering-online.com/content/12/1/66}.

\bibitem[Jamadandi et~al.(2018)Jamadandi, Kotturshettar, and
  Mudenagudi]{jamadandi2018predgan}
Adarsh Jamadandi, Sunidhi Kotturshettar, and Uma Mudenagudi.
\newblock Predgan: a deep multi-scale video prediction framework for detecting
  anomalies in videos.
\newblock In \emph{Proceedings of the 11th Indian Conference on Computer
  Vision, Graphics and Image Processing}, pages 1--8, 2018.

\bibitem[Khan(2016)]{Khan2016}
Shehroz Khan.
\newblock {Classification and Decision-Theoretic Framework for Detecting and
  Reporting Unseen Falls}. thesis.
\newblock Technical report, University of Waterloo. Ontario, Canada., 2016.

\bibitem[Khan and Hoey(2017)]{Khan2017}
Shehroz~S. Khan and Jesse Hoey.
\newblock {Review of fall detection techniques: A data availability
  perspective}.
\newblock \emph{Medical Engineering and Physics}, 39:\penalty0 12--22, 2017.
\newblock ISSN 18734030.
\newblock \doi{10.1016/j.medengphy.2016.10.014}.
\newblock URL \url{www.elsevier.com/locate/medengphy}.

\bibitem[Khan and Madden(2014)]{khan2014one}
Shehroz~S Khan and Michael~G Madden.
\newblock One-class classification: taxonomy of study and review of techniques.
\newblock \emph{The Knowledge Engineering Review}, 29\penalty0 (3):\penalty0
  345--374, 2014.

\bibitem[Khan et~al.(2021)Khan, Nogas, and Mihailidis]{khan2021spatio}
Shehroz~S Khan, Jacob Nogas, and Alex Mihailidis.
\newblock Spatio-temporal adversarial learning for detecting unseen falls.
\newblock \emph{Pattern Analysis and Applications}, 24\penalty0 (1):\penalty0
  381--391, 2021.

\bibitem[Kim et~al.(2022)Kim, Yu, Lee, and Kim]{kim2022video}
Yujun Kim, Jin-Yong Yu, Euijong Lee, and Young-Gab Kim.
\newblock Video anomaly detection using cross u-net and cascade sliding window.
\newblock \emph{Journal of King Saud University-Computer and Information
  Sciences}, 2022.

\bibitem[Kramarow(2015)]{KramarowEChenLHHedegaardH2015}
Ellen~A Kramarow.
\newblock \emph{Deaths from unintentional injury among adults aged 65 and over,
  United States, 2000-2013}.
\newblock Number 2015. US Department of Health and Human Services, Centers for
  Disease Control and~…, 2015.

\bibitem[Krizhevsky et~al.(2017)Krizhevsky, Sutskever, and
  Hinton]{krizhevsky2017imagenet}
Alex Krizhevsky, Ilya Sutskever, and Geoffrey~E Hinton.
\newblock Imagenet classification with deep convolutional neural networks.
\newblock \emph{Communications of the ACM}, 60\penalty0 (6):\penalty0 84--90,
  2017.

\bibitem[Liu et~al.(2018)Liu, Luo, Lian, and Gao]{liu2018future}
Wen Liu, Weixin Luo, Dongze Lian, and Shenghua Gao.
\newblock Future frame prediction for anomaly detection--a new baseline.
\newblock In \emph{Proceedings of the IEEE conference on computer vision and
  pattern recognition}, pages 6536--6545, 2018.

\bibitem[Mehta et~al.(2021)Mehta, Dhall, Pal, and Khan]{mehta2021motion}
Vineet Mehta, Abhinav Dhall, Sujata Pal, and Shehroz~S Khan.
\newblock Motion and region aware adversarial learning for fall detection with
  thermal imaging.
\newblock In \emph{2020 25th International Conference on Pattern Recognition
  (ICPR)}, pages 6321--6328. IEEE, 2021.

\bibitem[Milletari et~al.(2016)Milletari, Navab, and Ahmadi]{milletari2016v}
Fausto Milletari, Nassir Navab, and Seyed-Ahmad Ahmadi.
\newblock V-net: Fully convolutional neural networks for volumetric medical
  image segmentation.
\newblock In \emph{2016 fourth international conference on 3D vision (3DV)},
  pages 565--571. IEEE, 2016.

\bibitem[Mishra et~al.(2023)Mishra, Iaboni, Ye, Newman, Mihailidis, and
  Khan]{mishra2023privacy}
Pratik~K Mishra, Andrea Iaboni, Bing Ye, Kristine Newman, Alex Mihailidis, and
  Shehroz~S Khan.
\newblock Privacy-protecting behaviours of risk detection in people with
  dementia using videos.
\newblock \emph{BioMedical Engineering OnLine}, 22\penalty0 (1):\penalty0
  1--17, 2023.

\bibitem[Nguyen and Meunier(2019)]{nguyen2019anomaly}
Trong-Nguyen Nguyen and Jean Meunier.
\newblock Anomaly detection in video sequence with appearance-motion
  correspondence.
\newblock In \emph{Proceedings of the IEEE/CVF international conference on
  computer vision}, pages 1273--1283, 2019.

\bibitem[Nogas et~al.(2018)Nogas, Khan, and Mihailidis]{nogas2018fall}
Jacob Nogas, Shehroz~S. Khan, and Alex Mihailidis.
\newblock Fall detection from thermal camera using convolutional lstm
  autoencoder.
\newblock In \emph{Proceedings of the 2nd workshop on aging, rehabilitation and
  independent assisted living, IJCAI workshop}, 2018.

\bibitem[Nogas et~al.(2020)Nogas, Khan, and Mihailidis]{Nogas2020}
Jacob Nogas, Shehroz~S. Khan, and Alex Mihailidis.
\newblock {DeepFall: Non-Invasive Fall Detection with Deep Spatio-Temporal
  Convolutional Autoencoders}.
\newblock \emph{Journal of Healthcare Informatics Research}, 4\penalty0
  (1):\penalty0 50--70, mar 2020.
\newblock ISSN 2509498X.
\newblock \doi{10.1007/s41666-019-00061-4}.
\newblock URL
  \url{https://link.springer.com/article/10.1007/s41666-019-00061-4}.

\bibitem[Oktay et~al.(2018)Oktay, Schlemper, Folgoc, Lee, Heinrich, Misawa,
  Mori, McDonagh, Hammerla, Kainz, et~al.]{oktay2018attention}
Ozan Oktay, Jo~Schlemper, Loic~Le Folgoc, Matthew Lee, Mattias Heinrich,
  Kazunari Misawa, Kensaku Mori, Steven McDonagh, Nils~Y Hammerla, Bernhard
  Kainz, et~al.
\newblock Attention u-net: Learning where to look for the pancreas.
\newblock \emph{arXiv preprint arXiv:1804.03999}, 2018.

\bibitem[Ramachandran and Karuppiah(2020)]{ramachandran2020survey}
Anita Ramachandran and Anupama Karuppiah.
\newblock A survey on recent advances in wearable fall detection systems.
\newblock \emph{BioMed research international}, 2020, 2020.

\bibitem[Ronneberger et~al.(2015)Ronneberger, Fischer, and
  Brox]{ronneberger2015u}
Olaf Ronneberger, Philipp Fischer, and Thomas Brox.
\newblock U-net: Convolutional networks for biomedical image segmentation.
\newblock In \emph{International Conference on Medical image computing and
  computer-assisted intervention}, pages 234--241. Springer, 2015.

\bibitem[Rubenstein and Josephson(2002)]{Rubenstein2002}
Laurence~Z. Rubenstein and Karen~R. Josephson.
\newblock {The epidemiology of falls and syncope}.
\newblock \emph{Clinics in Geriatric Medicine}, 18\penalty0 (2):\penalty0
  141--158, 2002.
\newblock ISSN 07490690.
\newblock \doi{10.1016/S0749-0690(02)00002-2}.
\newblock URL \url{https://www.researchgate.net/publication/11207603}.

\bibitem[Schlemper et~al.(2019)Schlemper, Oktay, Schaap, Heinrich, Kainz,
  Glocker, and Rueckert]{schlemper2019attention}
Jo~Schlemper, Ozan Oktay, Michiel Schaap, Mattias Heinrich, Bernhard Kainz, Ben
  Glocker, and Daniel Rueckert.
\newblock Attention gated networks: Learning to leverage salient regions in
  medical images.
\newblock \emph{Medical image analysis}, 53:\penalty0 197--207, 2019.

\bibitem[Song et~al.(2015)Song, Lichtenberg, and Xiao]{song2015sun}
Shuran Song, Samuel~P Lichtenberg, and Jianxiong Xiao.
\newblock Sun rgb-d: A rgb-d scene understanding benchmark suite.
\newblock In \emph{Proceedings of the IEEE conference on computer vision and
  pattern recognition}, pages 567--576, 2015.

\bibitem[Stinchcombe et~al.(2014)Stinchcombe, Kuran, and
  Powell]{Stinchcombe2014}
A.~Stinchcombe, N.~Kuran, and S.~Powell.
\newblock {Seniors' falls in Canada: Second report: Key highlights}.
\newblock Technical Report 2-3, 2014.

\bibitem[Stone and Skubic(2014)]{stone2014fall}
Erik~E Stone and Marjorie Skubic.
\newblock Fall detection in homes of older adults using the microsoft kinect.
\newblock \emph{IEEE journal of biomedical and health informatics}, 19\penalty0
  (1):\penalty0 290--301, 2014.

\bibitem[Tang et~al.(2020)Tang, Zhao, Zhang, Gong, Li, and
  Yang]{tang2020integrating}
Yao Tang, Lin Zhao, Shanshan Zhang, Chen Gong, Guangyu Li, and Jian Yang.
\newblock Integrating prediction and reconstruction for anomaly detection.
\newblock \emph{Pattern Recognition Letters}, 129:\penalty0 123--130, 2020.

\bibitem[Vaswani et~al.(2017)Vaswani, Shazeer, Parmar, Uszkoreit, Jones, Gomez,
  Kaiser, and Polosukhin]{vaswani2017attention}
Ashish Vaswani, Noam Shazeer, Niki Parmar, Jakob Uszkoreit, Llion Jones,
  Aidan~N Gomez, {\L}ukasz Kaiser, and Illia Polosukhin.
\newblock Attention is all you need.
\newblock \emph{Advances in neural information processing systems}, 30, 2017.

\bibitem[Wu et~al.(2019)Wu, Liu, and Shen]{wu2019deep}
Peng Wu, Jing Liu, and Fang Shen.
\newblock A deep one-class neural network for anomalous event detection in
  complex scenes.
\newblock \emph{IEEE transactions on neural networks and learning systems},
  31\penalty0 (7):\penalty0 2609--2622, 2019.

\end{thebibliography}

\appendix

\end{document}